\documentclass[11pt]{article}

\usepackage[final]{coling}

\usepackage{times}
\usepackage{latexsym}

\usepackage[T1]{fontenc}

\usepackage{microtype}

\usepackage{inconsolata}

\usepackage{graphicx}

\usepackage{booktabs}
\usepackage{svg}
\usepackage{comment}
\usepackage{devanagari}
\usepackage{tikz}
\usepackage{caption}
\usepackage{subcaption}
\usepackage{hyperref}
\usepackage{tcolorbox}
\usepackage{xcolor}
\usepackage{amsmath}
\usepackage{amssymb}

\usepackage{arydshln}
\usepackage{multirow}
\usepackage{enumitem}
\usepackage{soul}
\newcommand{\rp}[1]{\textcolor{blue}{RP: #1}}
\newcommand{\mee}[1]{\textcolor{red}{#1}}

\title{MTQ-Eval: Multilingual Text Quality Evaluation for Language Models}

\author{Rhitabrat Pokharel \and Ameeta Agrawal\\
Department of Computer Science\\
Portland State University, USA \\
\texttt{\{pokharel,ameeta\}@pdx.edu} \\}

\begin{document}
\maketitle
\begin{abstract}
The use of large language models (LLMs) for evaluating outputs is becoming an increasingly effective and scalable approach. However, it remains uncertain whether this capability extends beyond task-specific evaluations to more general assessments of text quality, particularly in multilingual contexts. In this study, we introduce, MTQ-Eval, a novel framework for multilingual text quality evaluation {that learns from examples of both high- and low-quality texts, adjusting its internal representations.} To develop MTQ-Eval, we first automatically generate text quality preference data and then use it to train open-source base LLMs to align with ratings of high- and low-quality text. Our comprehensive evaluation across 115 languages demonstrates the improved performance of the proposed model. {Upon further analysis, we find that this enhanced evaluation capability also leads to notable improvements in downstream tasks.}
\end{abstract}

\section{Introduction}
Recent advances in LLMs have significantly enhanced their performance across various tasks, including text generation. While human evaluation remains the gold standard for assessing the quality of the generated text, it is costly, time-consuming, and particularly challenging to scale across languages. As a result, researchers have turned to LLMs themselves as automated evaluators for tasks such as translation, dialogue assessment, and essay scoring \cite{zheng2023judging,gilardi2023chatgpt,wangetal2023chatgpt,kocmifedermann2023large,ferronetal2023meep,stahletal2024exploring,nam2024using}. 

{Evaluating text quality is essential to ensure outputs are not only task-specifically accurate but also coherent, fluent, and natural to a native speaker.} However, there is still no standardized framework for assessing {general} quality of text using LLMs, especially in multilingual settings where linguistic diversity and data availability present significant challenges \cite{gala2023indictrans,bagheri-nezhad-agrawal-2024-drives}, and as LLM capabilities in multilingual contexts remain inconsistent \cite{hadaetal2024large,agrawal-etal-2024-evaluating}. 

\setlength{\tabcolsep}{3pt}
\begin{table*}[!t]
    \small
    \setlength{\tabcolsep}{4pt}
    \centering
    \begin{tabular}{p{12cm} p{1.1cm} p{1cm} p{1cm}}
    \toprule
    \textbf{Passage} &\textbf{GPT-4o} &\textbf{Llama-3.1} &\textbf{MTQ-Eval}\\
    \midrule
        \textbf{Original Text}: Red tide is caused by a higher than normal concentration of Karenia brevis, a naturally-occurring single-celled marine organism. Natural factors can intersect to produce ideal conditions, allowing this algae to increase in number dramatically. The algae produces a neurotoxin that can disable nerves in both humans and fish. Fish often die because of the high concentrations of the toxin in the waters. Humans can be affected by breathing affected water taken into the air by wind and waves. &4 &3 & 4\\
        \midrule
        \textbf{Word Shuffled}: Red tide is caused by a higher than normal concentration of Karenia brevis, a naturally-occurring single-celled marine organism. Natural factors can intersect \mee{humans} produce ideal conditions, allowing this algae to increase in number dramatically. The algae produces a neurotoxin that can disable nerves in both \mee{to} and fish. Fish often die because of the high concentrations of the toxin in the waters. Humans \mee{by} be affected \mee{can} breathing \mee{waves}. water taken into the air by \mee{and wind affected} &2 &2 & 1\\
        \bottomrule
    \end{tabular}
    \caption{Text quality ratings from three different models for the original and degraded text reveal that the scores for the original text are questionable for the \texttt{llama-3.1} model but these are improved when using MTQ-Eval.}
    \label{tab:deformed}
\end{table*}
 

\begin{figure*}[!t]
    \centering
    \includegraphics[width=1.0\textwidth]{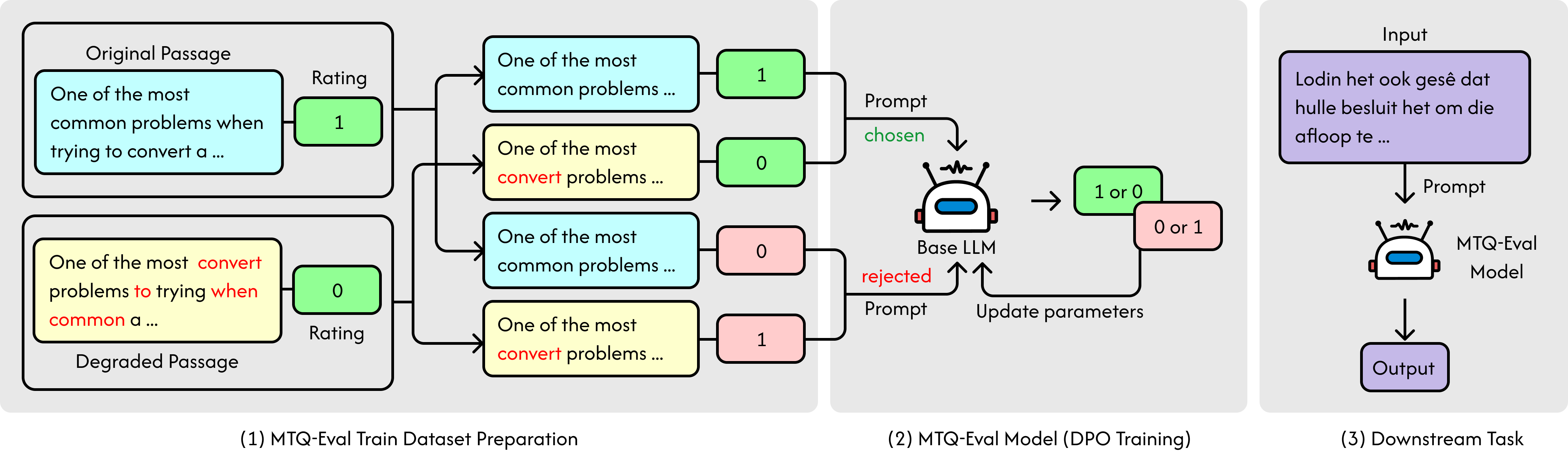}
    \caption{Overview of the MTQ-Eval (1) dataset creation, (2) model training, and (3) evaluation.}
    \label{fig:overview}
\end{figure*}


{Prompt-based text quality evaluation faces challenges in multilingual scenarios. For instance, while LLM-as-judge methods such as G-Eval \cite{liuetal2023g} can assess aspects like fluency and coherence, they are mainly effective in few-shot settings \cite{doostmohammadi-etal-2024-reliable}, which may not be practical for languages with limited data. Additionally, the scalability of prompt-based methods may depend on the model’s proficiency in understanding the target language. Furthermore, existing methods prioritize task-specific evaluation \cite{hadaetal2024large,huetal2024llm, bagheri-nezhad-etal-2025-beyond}  over general text quality. Complicating matters further, the concept of {text quality} remains broadly referenced but lacks a formal and universally accepted definition. In the literature, it has been discussed in various contexts such as natural language generation (NLG) quality \cite{huetal2024llm}, task quality, or content quality \cite{hadaetal2024large, hadaetal2024metal}.


Consider the example in Table \ref{tab:deformed} where LLMs such as GPT-4o and Llama 3.1 struggle {even in English} to assess the logical coherence of a passage  where the first passage of superior quality should have received a higher score (4 or 5 on a scale of 1 to 5). This raises a natural question: {\em Can a model reliably assess text quality across multiple languages, including low-resource languages?} This question leads to another compelling hypothesis: {\em Does a model's ability to effectively evaluate text quality translate to improved performance in other downstream tasks such as text classification or summarization?}  

To answer these questions, we introduce the Multilingual Text Quality Evaluation framework (MTQ-Eval), summarized in Figure~\ref{fig:overview}. Instead of relying on prompt-based evaluations that require language proficiency, we propose training a model to distinguish between high- and low-quality text. A major bottleneck in multilingual evaluation is the lack of good quality annotated datasets across languages. To overcome this, we develop an efficient and scalable synthetic data generation process that does not rely on human annotations, enabling the model to generalize across 115 languages. 


 Our work makes the following contributions: 
\begin{itemize}
    \item We propose a novel framework\footnote{The code is publicly available at \url{https://github.com/PortNLP/MTQ-Eval} and the models at \url{https://huggingface.co/PortNLP}} using Direct Preference Optimization (DPO) \cite{rafailov2023direct} to align models with text quality preference data, improving text evaluation in a wide range of languages.
    \item We automatically generate a dataset of high- and low-quality text samples across 115 languages without requiring human annotations.
\item We conduct extensive multilingual evaluations comparing our approach against baseline methods and also investigate whether proficiency in evaluating text quality maintains or improves {performance} in downstream tasks. 
\end{itemize}

\section{MTQ-Eval}
We consider text quality to encompass the following dimensions: {\bf coherence} (the logical flow and connectivity between sentences and ideas in the text as a whole), {\bf fluency} (the smoothness and grammatical accuracy), {\bf simplicity} (the ease of understanding), and {\bf linguistic acceptability} (the naturalness  and overall appropriateness), informed by prior research \cite{huetal2024llm, hadaetal2024large}.

\subsection{Dataset Creation} \label{sec:data}



To train LLMs to distinguish between high- and low-quality content across multiple languages, we need appropriate training data of preferences. 

\paragraph{\em High quality text} As base data, we leverage the high-quality text passages of  Belebele\footnote{\url{https://huggingface.co/datasets/facebook/belebele}}
dataset \cite{bandarkaretal2024belebele}, which is built upon Flores-200 \cite{nllb2022}. We select this dataset for its human-translated content in 100+ languages and a well-distributed spread of languages across both high and low resource levels, with a minimum of 43 languages represented in each level. High-resource (HR) languages are categorized under levels 3, 4, and 5, while low-resource (LR) languages correspond to levels 0, 1, and 2, as defined by \citet{joshietal2020state}. Besides, it is a parallel dataset that facilitates fair comparisons across languages.

\paragraph{\em Low quality text} Previous research has employed {perturbation} techniques such as reversing logical structures, altering verb inflections, and shuffling sentences. These methods are mostly feasible only for high-resource languages. Implementing them across more than hundred languages, most of them low-resource, is challenging due to the lack of robust NLP tools such as sentence tokenizers, connective detectors, and verb detectors.

Informed by prior work, we explore a controlled method for degrading text quality -- {word shuffling} \cite{kallini2024mission, huetal2024llm}. This straightforward technique, generally applicable across most languages, randomly rearranges a few words within a passage, disrupting the grammatical structure and semantic flow, creating noticeably distorted, low-quality text. {For word shuffling, we tokenized text by spaces and randomly swapped three to six words within each sample to introduce enough distortion. Too much distortion is also not good for model training  so that the degraded texts do not drift too far from the normal data.}  While useful, we acknowledge that this process may have varied impacts across languages, especially those with flexible word order. 

Out of the 122 languages in Belebele, we do not consider 7 languages {e.g., Basque and Japanese} {where word order plays lesser role and spaces do not function as reliable delimiters. {This results in a dataset of 115 languages.} From each language, we randomly selected 20 samples, yielding 2300 {\em normal} passage samples and 2300 {\em degraded} passages, for a total of 4,600 passages.

\paragraph{\em Quality annotations} As obtaining human annotations at scale for over 100 languages is infeasible, an alternative approach is to assign binary values to both normal and degraded text -- a score of 1 to the normal text, while a score of 0 to the corresponding degraded text (i.e.,  word shuffled text), as our model training (described in the next section) is formulated around relative ranking between two options, where one text is explicitly preferred over another. A small-scale manual review of the low-quality data across a handful of languages familiar to the authors confirmed its adequacy.

\subsection{Model Training}
To align the model with text quality judgments, we use Direct Preference Optimization (DPO) which has shown to be simpler and more stable than reinforcement learning from human feedback (RLHF) \cite{rafailov2023direct}. Moreover, unlike RLHF, DPO does not require explicit reward modeling, making it computationally efficient and easier to scale. {DPO has been successfully used in various model aligning tasks like multilingual alignment \cite{lai-etal-2023-okapi, wu-etal-2024-reuse, capo2025}, user intent alignment \cite{tunstall2023zephyr} and  diffusion model alignment \cite{wallace2024diffusion}. In this work, we explore this method to align a model to prefer high-quality text over low-quality text in multilingual settings.}


\begin{figure*}[!htbp]
\centering
\begin{tcolorbox}[colback=gray!10, colframe=gray, width=1.0\textwidth, rounded corners]
\small
\begin{verbatim}
Please act as an impartial judge and evaluate the text quality of the provided passage.
Text quality of a passage is defined by how well it maintains the following aspects.
(1) Coherence - logical flow and connectivity between sentences and ideas in the text.
(2) Fluency - smoothness and naturalness of individual sentences.
(3) Simplicity - how easy it is to understand the passage.
(4) Linguistic Acceptability - if the text sounds natural and correct to a native speaker.
Provide a binary rating, “0” for low quality or “1" for high quality, strictly following 
this format: “[[0]]” or “[[1]]“. Do not provide an explanation.
Passage: {passage}
\end{verbatim}
\end{tcolorbox}
\caption{Prompt used to obtain text quality ratings during DPO.}
\label{fig:prompt_template}

\end{figure*}

DPO fine-tunes the model by increasing the likelihood of preferred responses while reducing the probability of less preferred ones. The model directly optimizes for relative preference ranking through a simple classification objective. The optimization follows a log-ratio loss function, which is defined as:
\[
L_{\text{DPO}} = - \log \frac{\exp(\pi_\theta(x_{\text{good}}))}{\exp(\pi_\theta(x_{\text{good}})) + \exp(\pi_\theta(x_{\text{bad}}))}
\]

\noindent where \( x_{\text{good}} \) represents a high-quality text sample, \( x_{\text{bad}} \) is a low-quality text sample, and \( \pi_\theta \) denotes the policy network (the LLM being fine-tuned).

This loss formulation encourages the model to rank the better text higher rather than simply classifying quality in isolation. By training on direct comparisons rather than absolute scores, DPO provides a robust framework for learning text preferences across diverse languages and linguistic structures.


A DPO method requires three key components from the dataset: 
\begin{itemize}[itemsep=0pt]
    \item \emph{prompt} (the instruction given to the model),
    \item \emph{chosen response} (the preferred high-quality text sample), and
    \item \emph{rejected response} (the less preferred low-quality text sample).
\end{itemize}  

To construct \emph{prompts}, we ask the model to evaluate and rate a given passage. The {prompt} template is based on a combination of methodologies from {G-Eval} \cite{liuetal2023g} and {LLM-as-a-Judge} \cite{zheng2023judging}, and is illustrated in Figure \ref{fig:prompt_template}}. For the \emph{chosen} response, we provide the original passage along with its expected rating, while in the \emph{rejected} response, we provide the same passage but with an incorrect rating. For instance, for original high-quality passages, a chosen rating is 1 and a rejected rating is 0. Similarly, for low-quality passages, we assign a chosen rating of 0 and a rejected rating of 1. This ensures a clear contrast between the ratings of normal passages and degraded texts. A simplified version of the data preparation process is illustrated in Figure~\ref{fig:training_data_format} with detailed examples of each component provided in Appendix \ref{sec:dpo_dataset_format}.

\begin{figure}[!t]
    \centering
    \includegraphics[width=0.35\textwidth]{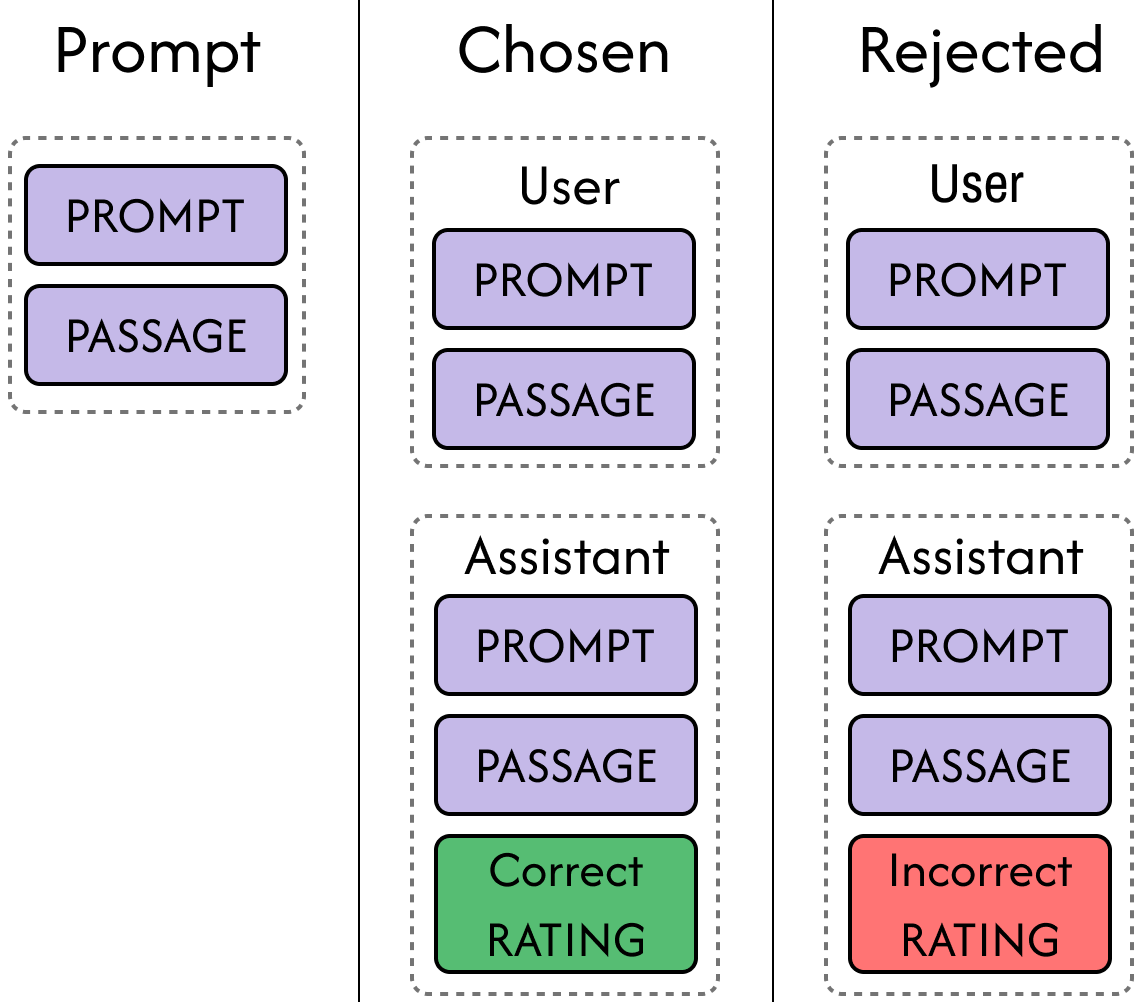}
    \caption{The training dataset format for DPO training.}\vspace{-0.3cm}
\label{fig:training_data_format}
\end{figure}

    






\section{Evaluation Setup}
We conduct a comprehensive evaluation of our proposed model across various experimental settings {as shown in Table~\ref{tab:exp_setup}}. This section details the models, datasets, and evaluation metrics used in our study. 

\begin{table}[!t]
\centering
\small
\begin{tabular}{p{2.2cm} p{2cm} p{2.8cm}}
\toprule
\textbf{Task} & \textbf{Train Data} & \textbf{Model} \\
\midrule
\textbf{\underline{Text-quality}} \\
Belebele & Train split & SFT model for Belebele \\
Belebele & Train split & Using DPO to create MTQ-Eval \\
MELA & Train split & SFT model for MELA \\
\midrule
\textbf{\underline{Downstream}} \\
Sentiment Analysis (MMS) & Train split & SFT model for MMS \\
Summarization (XLSum) & Train split & SFT model for XLSum \\
\bottomrule
\end{tabular}
\caption{Overview of tasks, corresponding training data, and models used in our experiments.}
\label{tab:exp_setup}
\end{table}



\subsection{Models and Baselines}
We use Llama 3.1 Instruct 8B\footnote{\url{https://huggingface.co/meta-llama/Meta-Llama-3.1-8B-Instruct}} (\texttt{llama}) and Aya Expanse 8B\footnote{\url{https://huggingface.co/CohereForAI/aya-expanse-8b}} (\texttt{aya}) as base models due to their extensive multilingual support: \texttt{llama} supports 8 languages, while \texttt{aya} performs well in 23 languages. During experiments, we compare MTQ-Eval (\texttt{llama} and \texttt{aya}) against base models using prompt-based inference and supervised fine-tuned (SFT) versions of these models. We train multiple SFT models specific to each task/dataset using 20 labeled samples per language from the corresponding dataset. For fine-tuning all models, we apply LoRA \cite{hu2022lora} with these configurations -- alpha value of 128, dropout rate of 0.05, rank of 64, and one epoch of training with a batch size of 2 and a learning rate of 5e-7. 


\subsection{Datasets and Metrics}
We evaluate our models using two text quality evaluation datasets and two downstream task datasets. 

\subsubsection{Text Quality Evaluation}
 
\paragraph{\em Linguistic Acceptability}  We use the  Multilingual Evaluation of Linguistic Acceptability (MELA) dataset \cite{zhangetal2024mela}, which provides human ratings for linguistic acceptability across 10 languages (9 high-resource, 1 low-resource). This dataset assesses whether a sentence is syntactically acceptable to native speakers. We use the original  CoLA-like prompt \cite{warstadtetal2019neural} (detailed in Appendix \ref{sec:prompt_dimension_specific}). We randomly sample 20 examples per language (total 200 samples)  and report the results in terms of Matthew’s Correlation Coefficient (MCC).

\paragraph{\em Text Quality Evaluation using Belebele Passages} \quad While MELA provides human ratings, it is limited to only 10 languages. Scaling such judgments to 100+ languages is infeasible, so we construct an alternative test dataset using the methodology described in Section~\ref{sec:data}. We extract a different set of passages from Belebele and create a test set of 2,300 high-quality samples and 2,300 low-quality samples (20 samples each across 115 languages) resulting in a total of 4,600 test instances. The distribution includes 43 HR languages, 61 LR languages, and a few  uncategorizable ones. {This dataset also covers a diverse range of language families and scripts.} During inference, we use the same prompt shown in Figure \ref{fig:prompt_template}, with the addition of \texttt{Your Answer:} at the end. We report results using MCC along with Kullback-Leibler (KL) divergence which measures how distinguishable the ratings of normal text are compared to degraded text, as well as F1 scores.




\subsection{Downstream Task Evaluation Datasets and Metrics}
To assess whether improvements in text quality evaluation enhance performance in other NLP tasks, we evaluate models on multilingual text classification and text generation.

\paragraph{\em Sentiment Analysis}  
We use the Massively Multilingual Corpus of Sentiment (MMS) dataset \cite{augustyniak2023massively} which includes 28 languages spanning six language families. The dataset is labeled with sentiment labels: positive, neutral, and negative. We select 20 samples per language, resulting in a total of 560 samples and report the results in terms of F1 scores. 
The prompt is based on LLM-as-a-Judge concept \cite{zheng2023judging} (details in Appendix~\ref{sec:prompt_dimension_specific}).

\paragraph{\em Summarization} We use the XL-Sum dataset \cite{hasan-etal-2021-xl} which contains  article-summary pairs sourced from the BBC across 45 languages, ranging from low to high resource. We sample 20 articles from each language (total 900 samples) and generate summaries using a prompt adapted from \citep{palm_2023} (details in Appendix~\ref{sec:prompt_dimension_specific}). We evaluate the results using  {G-Eval} \cite{liuetal2023g}  across three dimensions: coherence, consistency, and fluency. 

\section{Results of Text Quality Evaluation}

In this section, we discuss the results of assessing multilingual text quality.

\begin{table}[!t]
\centering
\setlength{\tabcolsep}{8pt}
\small
\begin{tabular}{lcc|ccc}
\toprule
& \multicolumn{2}{c}{\textbf{\texttt{llama}}} & \multicolumn{2}{c}{\textbf{\texttt{aya}}}\\
\textbf{Lang.} & \textbf{SFT} & \textbf{MTQ-Eval} & \textbf{SFT} & \textbf{MTQ-Eval} \\
\midrule
ara &\textbf{0.61} &0.00 &0.31 &\textbf{0.73} \\
deu &\textbf{0.00} &-0.25 &\textbf{0.52} &0.31 \\
eng &0.23 &\textbf{0.35} &0.50 &\textbf{0.52} \\
spa &0.00 &\textbf{0.42} &0.23 &\textbf{0.25} \\
fra &0.14 &\textbf{0.25} &\textbf{0.31} &0.22 \\
isl &\textbf{0.00} &-0.23 &-0.23 &\textbf{0.33} \\
ita &\textbf{0.33} &0.23 &0.00 &\textbf{0.42} \\
jpn &-0.14 &\textbf{0.25} &\textbf{0.23} &0.20 \\
rus &0.23 &\textbf{0.33} &0.23 &\textbf{0.42} \\
zho &0.33 &\textbf{0.42} &0.00 &\textbf{0.50} \\
\midrule
\textbf{Avg.} &0.17 &\textbf{0.18} &0.21 &\textbf{0.39} \\
\bottomrule
\end{tabular}
\caption{MCC results on MELA dataset.}
\label{tab:mela_results}
\end{table}

\begin{table}[!t]
\centering
\setlength{\tabcolsep}{8pt}
\small
\begin{tabular}{lccc}
\toprule
&\textbf{MCC $\uparrow$} &\textbf{KL Div. $\uparrow$} &\textbf{F1 $\uparrow$}\\
\midrule
\texttt{llama} base &0.18&0.06  & 0.56\\
\texttt{llama} SFT   &0.17&0.07 & 0.57\\
\texttt{llama} MTQ-Eval   &\textbf{0.24}&\textbf{0.12} & \textbf{0.59}\\
\midrule
\texttt{aya} base  &0.14&0.04 &0.44 \\
\texttt{aya} SFT   &-0.23&0.10 &0.35 \\
\texttt{aya} MTQ-Eval  &\textbf{0.26}&\textbf{0.15} & \textbf{0.60} \\
\bottomrule
\end{tabular}
\caption{MCC, KL Divergence, and F1 scores on {Belebele text quality evaluation dataset}.}
\label{tab:diff_metric_results}
\end{table}

\subsection{Performance on \texttt{MELA} dataset}
Table~\ref{tab:mela_results} presents the MCC scores for text quality evaluation on the MELA dataset across ten languages using both \texttt{llama} and \texttt{aya} models, evaluated under SFT and MTQ-Eval paradigms. Across both models, the MTQ-Eval approach generally improves performance over SFT, particularly for \texttt{aya}, which achieves an average MCC of 0.39-- a substantial improvement over SFT (0.21). \texttt{llama} sees a smaller but still positive gain.

For individual languages,  \texttt{llama} MTQ-Eval is particularly beneficial for Spanish and Chinese, while \texttt{aya} MTQ-Eval shows strong gains in Arabic and Chinese. Interestingly, even English benefits from MTQ-Eval in both models. Overall, these results suggest that MTQ-Eval enhances multilingual text quality evaluation across most languages.


\subsection{Performance on \texttt{Belebele} text quality dataset}

Encouraged by these results, we extend our evaluation to a test set created using the Belebele dataset. From the average results covering 115 languages presented in Table~\ref{tab:diff_metric_results}, we observe that across both models, MTQ-Eval consistently outperforms both the base and SFT versions. The MCC scores show that MTQ-Eval correlates better with expected ratings (with an improvement in MCC from 0.18 to 0.24 for \texttt{llama} and 0.14 to 0.26 for aya), while the KL divergence results further confirm {MTQ-Eval}'s effectiveness in distinguishing between high- and low-quality text. The stark underperformance of \texttt{aya} SFT highlights the  robustness of preference-based training over standard supervised fine-tuning. The F1 score results (the ground truth for normal text is 1, and for degraded text, it is 0),  also support that the MTQ-Eval models are effectively distinguishing between normal and degraded text.




\paragraph{\em High- vs. Low-Resource Languages} Regarding performance for high- and low-resource languages, the MCC scores in Table~\ref{tab:mcc_hr_lr } show that  MTQ-Eval generally shows improved performance for both models, particularly in high-resource languages. The gains in low-resource languages are less pronounced, reflecting the general challenge of low-resource modeling.

A further analysis of these results across a more granular breakdown of resource levels (0-5) is shown in Table~\ref{tab:f1_by_rl}. In low-resource settings (levels 0-2), \texttt{llama} MTQ-Eval achieves an average F1 score of 0.55, slightly outperforming \texttt{aya} (0.50). However, the gap widens in high-resource settings (levels 3-5) with \texttt{aya} significantly outperforming llama, achieving an average F1 score of 0.77 as compared to 0.65 for llama. This indicates that \texttt{aya} benefits more from MTQ-Eval for higher-resource languages.


\begin{table}[!t]
\centering
\setlength{\tabcolsep}{8pt}
\small
\begin{tabular}{lcc|cc}
\toprule
& \multicolumn{2}{c}{\textbf{MCC} $\uparrow$} & \multicolumn{2}{c}{\textbf{F1} $\uparrow$} \\
& \textbf{LR} & \textbf{HR} & \textbf{LR} & \textbf{HR} \\
\midrule
\texttt{llama} base & 0.11 & 0.29 & 0.52 & 0.61 \\
\texttt{llama} SFT & 0.11 & 0.29 & 0.54 & 0.63 \\
\texttt{llama} MTQ-Eval & \textbf{0.16} & \textbf{0.38} & \textbf{0.55} & \textbf{0.64} \\
\midrule
\texttt{aya} base & \textbf{0.12} & 0.19 & 0.43 & 0.42 \\
\texttt{aya} SFT & -0.06 & -0.43 & 0.42 & 0.28 \\
\texttt{aya} MTQ-Eval & 0.07 & \textbf{0.50} & \textbf{0.50} & \textbf{0.75} \\
\bottomrule
\end{tabular}
\caption{MCC and F1 scores on Belebele text quality evaluation dataset for high-resource (HR) and low-resource (LR) languages.}
\label{tab:mcc_hr_lr }
\end{table}

\begin{table}[!t]
\centering
\setlength{\tabcolsep}{10pt}
\small
\begin{tabular}{lcc}
\toprule
\textbf{Res. Level} & \textbf{MTQ-Eval} &\textbf{MTQ-Eval}\\
& \texttt{llama}  &  \texttt{aya}\\
\midrule
0 &0.58 &0.50 \\
1 &0.55 &0.51 \\
2 &0.52 &0.49 \\
\midrule
LR Avg. & 0.55 &0.50\\
\midrule
3 &0.61 &0.69 \\
4 &0.68 &0.82 \\
5 &0.67 &0.82 \\
\midrule
HR Avg. &0.65 &0.77\\
\bottomrule
\end{tabular}
\caption{F1 scores by resource levels on Belebele text quality evaluation dataset.}
\label{tab:f1_by_rl}
\end{table}

\begin{figure}[!t]
    \centering
    \begin{subfigure}[b]{0.23\textwidth}
        \centering
        \includegraphics[width=\textwidth]{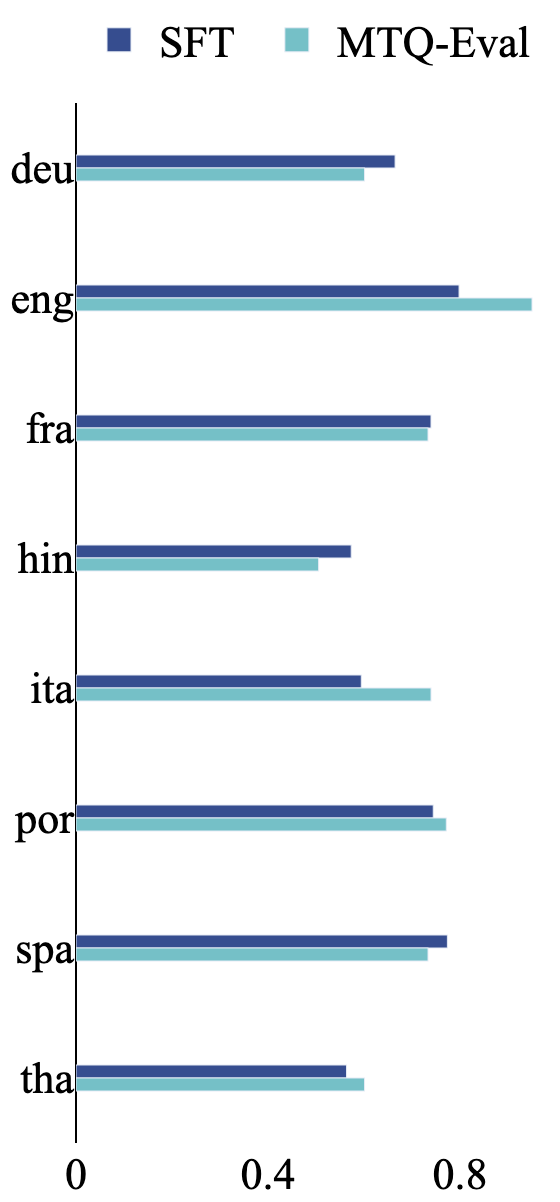}
        \caption{ \texttt{llama}}
        \label{fig:f1_llama_preferred}
    \end{subfigure}
    \begin{subfigure}[b]{0.23\textwidth}
        \centering
        \includegraphics[width=\textwidth]{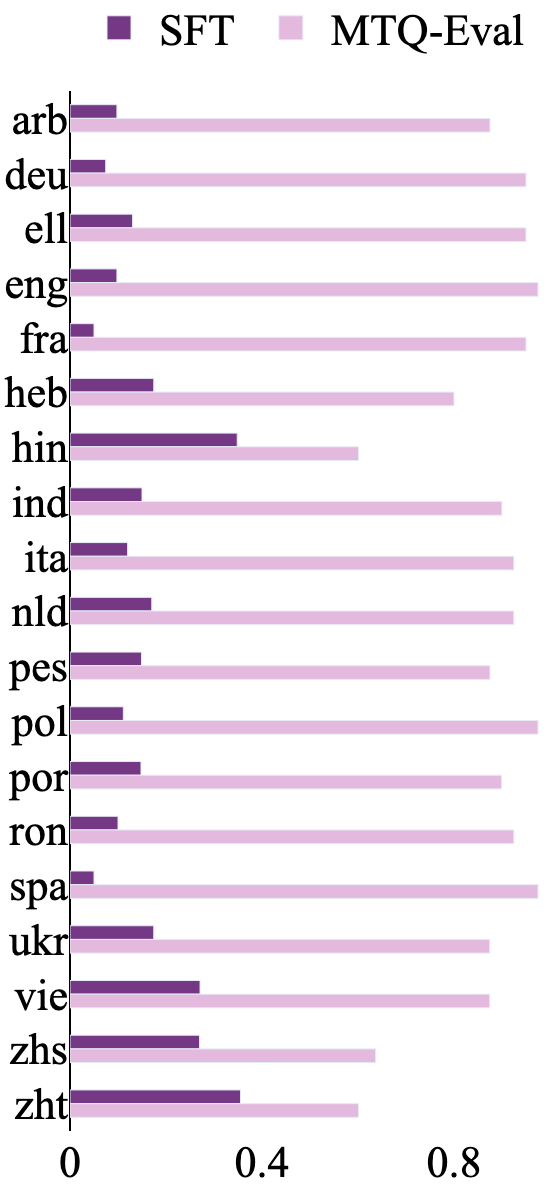}
        \caption{\texttt{aya}}
        \label{fig:f1_ayaexp_preferred}
    \end{subfigure}
\caption{F1 scores of supported languages}
\label{fig:supported}
\end{figure}

\begin{table}[!t]
\centering
\setlength{\tabcolsep}{3.5pt}
\small
\begin{tabular}{lc|lc}
\toprule
\textbf{Lang. Family} & \textbf{MTQ-Eval} &\textbf{Lang. Family} &\textbf{MTQ-Eval}\\
& \texttt{llama}  & & \texttt{aya}\\
\midrule
Creole &0.67 &Indo-European &0.72 \\
Austronesian &0.65 &Austro-Asiatic &0.70 \\
Turkic &0.63 &Kartvelian &0.61 \\
Indo-European &0.62 &Afro-Asiatic &0.59 \\
Uralic &0.61 &Sino-Tibetan &0.57 \\
Indo-Aryan &0.60 &Austronesian &0.56 \\
Afro-Asiatic &0.57 &Dravidian &0.53 \\
Tupian &0.55 &Uralic &0.52 \\
Niger-Congo &0.54 &Kra-Dai &0.51 \\
Austro-Asiatic &0.54 &Turkic &0.48 \\
Sino-Tibetan &0.53 &Mongolic &0.46 \\
Mongolic &0.52 &Niger-Congo &0.34 \\
Kra-Dai &0.52 &Tupian &0.33 \\
Kartvelian &0.39 &Creole &0.33 \\
Nilo-Saharan &0.37 &Indo-Aryan &0.33 \\
Dravidian &0.36 &Nilo-Saharan &0.33 \\
\bottomrule
\end{tabular}
\caption{F1 scores across language families on Belebele text quality evaluation dataset.}\vspace{-0.4cm}
\label{tab:f1_lang_family}
\end{table}

\begin{table*}[!t]\centering
\setlength{\tabcolsep}{6pt}
\small
\begin{tabular}{lclc|lclc}
\toprule
\multicolumn{4}{c}{\textbf{MTQ-Eval \texttt{llama}}} & \multicolumn{4}{c}{\textbf{MTQ-Eval \texttt{aya}}}\\
\textbf{Language} &\textbf{F1 (Top 10)} &\textbf{Language} &\textbf{F1 (Bottom 10)} &\textbf{Language} &\textbf{F1 (Top 10)} &\textbf{Language} &\textbf{F1 (Bottom 10)} \\
\midrule
eng\_Latn &0.95 &mal\_Mlym &0.39 &spa\_Latn &0.97 &kac\_Latn &0.33 \\
nld\_Latn &0.80 &khm\_Khmr &0.39 &pol\_Latn &0.97 &sot\_Latn &0.33 \\
ron\_Latn &0.80 &kan\_Knda &0.37  &eng\_Latn &0.97 &sna\_Latn &0.33 \\
nob\_Latn &0.80 &bod\_Tibt &0.37 &fra\_Latn &0.95 &kin\_Latn &0.33 \\
pes\_Arab &0.77 &amh\_Ethi &0.37 &ell\_Grek &0.95 &sin\_Sinh &0.33 \\
por\_Latn &0.77 &asm\_Beng &0.37 &deu\_Latn &0.95 &lin\_Latn &0.33 \\
swe\_Latn &0.76 &tel\_Telu &0.36  &ces\_Latn &0.92 &lug\_Latn &0.33 \\
nya\_Latn &0.75 &ben\_Beng &0.36 &ron\_Latn &0.92 &nso\_Latn &0.33 \\
zsm\_Latn &0.74&ory\_Orya &0.33  &ita\_Latn &0.92 &pbt\_Arab &0.33 \\
ita\_Latn &0.74 &tam\_Taml &0.33  &nld\_Latn &0.92 &zul\_Latn &0.33 \\
\bottomrule
\end{tabular}
\caption{Top and bottom performing languages on Belebele text quality evaluation dataset.}
\label{tab:f1_top_bottom_llama_ayaexp}
\end{table*}

\begin{table}[!t]
\centering
\setlength{\tabcolsep}{4pt}
\small
\begin{tabular}{lc|lc}
\toprule
\multicolumn{2}{c}{\textbf{MTQ-Eval \texttt{llama}}} & \multicolumn{2}{c}{\textbf{MTQ-Eval \texttt{aya}}}\\
\textbf{Script} &\textbf{\% Improvement} &\textbf{Script} &\textbf{\% Improvement} \\\midrule
Khmr &30 &Hebr &105 \\
Mlym &18 &Hans &94 \\
Tibt &16 &Taml &94 \\
Deva &15 &Geor &85 \\
Ethi &15 &Grek &83 \\
Thai &15 &Hant &82 \\
Cyrl &14 &Telu &70 \\
Hans &13 &Arab &65 \\
Telu &12 &Mymr &64 \\
Mymr &9 &Gujr &58 \\
Hebr &9 &Armn &58 \\
Laoo &8 &Cyrl &57 \\
Knda &6 &Guru &55 \\
Arab &5 &Deva &54 \\
Latn &5 &Beng &49 \\
Guru &0 &Thai &48 \\
Gujr &-2 &Khmr &39 \\
Sinh &-2 &Mlym &39 \\
Hant &-6 &Laoo &33 \\
Grek &-6 &Tibt &30 \\
Geor &-11 &Knda &21 \\
Armn &-14 &Ethi &21 \\
Orya &-21 &Latn &20 \\
Taml &-21 &Orya &0 \\
Beng &-26 &Sinh &0 \\
\bottomrule
\end{tabular}
\caption{Improvement in F1 score (\%) by MTQ-Eval over base model for scripts in Belebele dataset.}
\label{tab:scripts_llama3_ayaexp}
\end{table}

\paragraph{\em Supported Languages} Our next analysis specifically focuses on languages supported by \texttt{llama} (8 languages) and \texttt{aya} (23 languages). Figure~\ref{fig:supported} shows that for half of the model-supported languages, MTQ-Eval outperforms SFT under \texttt{llama}. However, \texttt{aya} shows a much stronger contrast between MTQ-Eval and SFT, with MTQ-Eval showing substantial improvements in all languages. 




\paragraph{\em Language Families} Next, we examine the performance across language families. The results presented in {Table \ref{tab:f1_lang_family} show that \texttt{llama} MTQ-Eval performs best in Creole and Austronesian languages, while \texttt{aya} MTQ-Eval does well in Indo-European and Austro-Asiatic languages. Both models struggle with Nilo-Saharan languages, indicating a common weakness in underrepresented linguistic groups.

Table~\ref{tab:f1_top_bottom_llama_ayaexp} presents the top and bottom performing languages (full results in Appendix~\ref{app:top}). For \texttt{llama} MTQ-Eval, the top 20 performing languages included both high-resource (60\%) and low-resource (40\%) languages, with 65\% being Indo-European and 95\% using the Latin script. In contrast, the bottom 20 consisted of 80\% low-resource languages, with several diverse scripts. For \texttt{aya} MTQ-Eval, the top 20 was exclusively high-resource languages, with 80\% Indo-European and 75\% Latin script. The bottom 20 was dominated by low-resource languages, 65\% of which were Niger-Congo, and 85\% used the Latin script. Interestingly, right-to-left script languages (Persian, Arabic)  show strong performance in both models. Overall, both models perform well in high-resource European languages, while showing weaknesses in South Asian scripts and African languages.

\paragraph{\em Scripts} Table~\ref{tab:scripts_llama3_ayaexp} presents the percentage improvement by script with MTQ-Eval over base models. While \texttt{llama} MTQ-Eval benefits Khmer, Malayalam, and {Tibetan} scripts, \texttt{aya} MTQ-Eval shows the largest improvement for {Hebrew, Simplified Chinese (Hans), and Tamil} scripts. However, \texttt{llama} MTQ-Eval negatively impacts some scripts (Orya, Tamil, and Bengali). Overall, \texttt{aya} benefits more from MTQ-eval across a wider range of scripts as compared to \texttt{llama}.

\begin{table}[!t]
\centering
\setlength{\tabcolsep}{6pt}
\small
\begin{tabular}{lcc|ccc}
\toprule
& \multicolumn{2}{c}{\textbf{\texttt{llama}}} & \multicolumn{2}{c}{\textbf{\texttt{aya}}}\\
\textbf{Lang.} & \textbf{SFT} & \textbf{MTQ-Eval} & \textbf{SFT} & \textbf{MTQ-Eval} \\
\midrule
ara &0.55 &\textbf{0.71} &\textbf{0.67} &\textbf{0.67} \\
bul &\textbf{0.23} &0.13 &0.19 &\textbf{0.32} \\
bos &0.44 &\textbf{0.62} &0.48 &\textbf{0.62} \\
cze &\textbf{0.54} &0.28 &\textbf{0.62} &0.47 \\
deu &\textbf{0.44} &\textbf{0.44} &\textbf{0.41} &0.29 \\
ell &0.19 &\textbf{0.37} &\textbf{0.35} &0.21 \\
eng &\textbf{0.80} &0.75 &\textbf{0.80} &0.75 \\
spa &0.58 &\textbf{0.69} &0.31 &\textbf{0.52} \\
fas &0.47 &\textbf{0.55} &0.64 &\textbf{0.75} \\
fra &\textbf{0.71} &0.49 &\textbf{0.68} &0.64 \\
heb &\textbf{0.40} &\textbf{0.40} &\textbf{0.47} &0.45 \\
hin &0.42 &\textbf{0.46} &\textbf{0.61} &0.42 \\
hrv &0.38 &\textbf{0.46} &0.16 &\textbf{0.46} \\
hun &0.38 &\textbf{0.5} &\textbf{0.48} &0.42 \\
ita &0.43 &\textbf{0.47} &0.24 &\textbf{0.45} \\
jpn &0.52 &\textbf{0.75} &\textbf{0.71} &0.61 \\
lav &\textbf{0.43} &0.39 &0.53 &\textbf{0.54} \\
pol &\textbf{0.56} &\textbf{0.56} &0.77 &\textbf{0.80} \\
por &\textbf{0.52} &0.49 &\textbf{0.60} &\textbf{0.60} \\
rus &0.52 &\textbf{0.6} &\textbf{0.53} &0.44 \\
slk &0.55 &\textbf{0.56} &0.28 &\textbf{0.57} \\
slv &0.40 &\textbf{0.45} &0.29 &\textbf{0.49} \\
sqi &\textbf{0.26} &0.23 &\textbf{0.40} &0.23 \\
srp &\textbf{0.56} &0.54 &0.19 &\textbf{0.50} \\
swe &0.40 &\textbf{0.51} &0.40 &\textbf{0.41} \\
tha &0.58 &\textbf{0.68} &0.48 &\textbf{0.58} \\
urd &\textbf{0.57} &0.53 &0.48 &\textbf{0.71} \\
zho &\textbf{0.75} &0.64 &0.66 &\textbf{0.74} \\
\midrule
\textbf{Avg} &0.49 &\textbf{0.51} &0.48 &\textbf{0.52} \\
\bottomrule
\end{tabular}
\caption{F1 scores on the MMS dataset.}
\label{tab:mms_results}
\end{table}


\begin{table*}[!t]
\centering
\setlength{\tabcolsep}{6pt}
\small
\begin{tabular}{lrrrr}\toprule
\textbf{Model} & \textbf{Coh.} & \textbf{Con.} & \textbf{Flu.} & \textbf{Avg.}\\
\midrule
\texttt{llama} SFT (LR) & 3.25 $\pm$ 0.04 & 3.10 $\pm$ 0.11 & 2.99 $\pm$ 0.01 & 3.11\\
\texttt{llama} MTQ-Eval (LR) & 3.25 $\pm$ 0.04 & 3.18 $\pm$ 0.10 & 2.99 $\pm$ 0.01 & \textbf{3.14}\\
\hdashline
\texttt{llama} SFT (HR) & 3.22 $\pm$ 0.04 & 3.18 $\pm$ 0.11 & 2.98 $\pm$ 0.01 & \textbf{3.13}\\
\texttt{llama} MTQ-Eval (HR) & 3.20 $\pm$ 0.04 & 3.17 $\pm$ 0.11 & 2.98 $\pm$ 0.01 & 3.12\\
\midrule
\texttt{aya} SFT (LR) & 3.24 $\pm$ 0.04 & 3.09 $\pm$ 0.12 & 2.97 $\pm$ 0.02 & 3.10\\
\texttt{aya} MTQ-Eval (LR) & 3.22 $\pm$ 0.04 & 3.13 $\pm$ 0.12 & 2.98 $\pm$ 0.01 & \textbf{3.11}\\
\hdashline
\texttt{aya} SFT (HR) & 3.22 $\pm$ 0.04 & 3.07 $\pm$ 0.11 & 2.98 $\pm$ 0.01 & \textbf{3.09}\\
\texttt{aya} MTQ-Eval (HR) & 3.24 $\pm$ 0.04 & 3.01 $\pm$ 0.11 & 2.98 $\pm$ 0.01 & 3.08\\
\bottomrule
\end{tabular}
\caption{Mean scores with 95\% confidence intervals (± CI) for XLSum summaries. Confidence intervals were computed using the t-distribution and remain small and consistent across models.}
\label{tab:mean_sd_llama_aya}
\end{table*}



\section{Downstream Task Performance}
Having verified  MTQ-Eval's effectiveness in text quality assessment, we now evaluate whether its benefits extend to downstream tasks like sentiment analysis and summarization.

\subsection{Sentiment Analysis}
Table~\ref{tab:mms_results} shows the F1 scores on the MMS dataset. When trained on \texttt{llama}, the average F1 score of 0.51 with MTQ-Eval and 0.49 with SFT demonstrates that our method performs slightly better. \texttt{aya} MTQ-Eval sees similar improvements with average F1 score of 0.52, compared to 0.48 with SFT. Recall that MTQ-Eval is trained on synthetic text quality data while SFT is trained on task-specific labeled data.  While the gains are modest, the results suggest that models trained to assess text quality also develop stronger representations, which help with classification tasks. More importantly, a key advantage of our method is that it does not require training with {\em task-specific} labeled data. Instead, with only a small set of synthetically generated text quality samples, it can boost downstream performance. This is particularly valuable in multilingual settings, where obtaining sufficient training data for low-resource languages is often challenging.


\subsection{Summarization}
Table~\ref{tab:mean_sd_llama_aya} presents the mean scores of evaluating summary quality across high- and low-resource languages of XL-Sum dataset. We observe that MTQ-Eval performs as well as or better than SFT across all three evaluation dimensions, and particularly in low-resource categories ({(+0.03 in Llama, +0.01 in Aya-Expanse)}). This demonstrates that MTQ-Eval naturally enhances the ability to generate high-quality content, without explicit training on task-specific data.

{We also analyzed English (HR) and Nepali (LR) summaries to explore behavior across language settings in Table~\ref{tab:summary_examples}.  MTQ-Eval shows only slight improvements in English, where SFT already generates strong summaries. In contrast, for Nepali, where SFT outputs are noisy, MTQ-Eval produces coherent summaries. This highlights its greater impact in LR languages compared to HR ones.}

\begin{table*}[!t]
    \centering
    \includegraphics[width=1.0\textwidth]{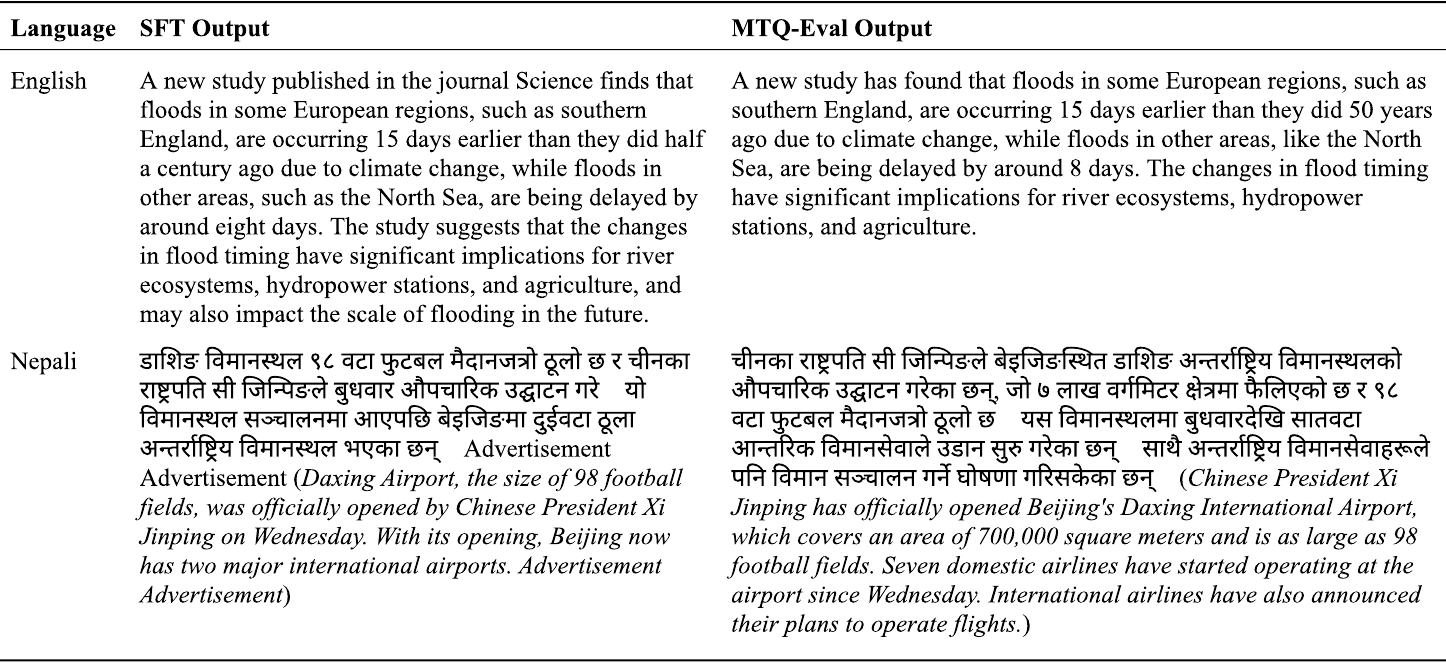}
    \caption{Summaries in English and Nepali generated by SFT and MTQ-Eval models for a news article.}
    \vspace{-0.3cm}
\label{tab:summary_examples}
\end{table*}

\section{Related Work}
\textbf{\em LLM-as-an-Evaluator} \quad The use of LLMs to evaluate their own outputs has recently gained significant attention. \citet{zheng2023judging} compared LLMs’ judgments with human evaluations on tasks like multi-turn dialogues and the Chatbot Arena. 
\citet{huetal2024llm} used perturbations to explore various aspects of text quality using prompt-based methods, but found that such methods often do not consistently provide reliable scores. Similarly, \citet{huang2024empirical} highlighted the limitations of fine-tuned judge models for LLM evaluation, noting that these models are typically task-specific and lack generalizability, while others questioned LLMs' potential to replace human evaluators \cite{chianglee2023large}. 

Several prompt-based approaches have also been proposed  although the findings have been mixed. For example, \citet{huetal2024llm} showed that varying the details in the evaluation criteria had little effect on LLM behavior. {\citet{murugadoss2024evaluating} found that detailed instructions are not always helpful in improving automatic evaluations made by LLMs. \citet{doostmohammadi-etal-2024-reliable} noticed that prompt based method works best mostly under few shot setting.} In contrast, \citet{hadaetal2024metal, lee2024checkeval} demonstrated that detailed instructions resulted in evaluations that closely resembled human judgments. {\citet{chollampatt-etal-2025-cross} leveraged machine translation and a reference LLM to generate a reference for comparison in high resource language.} To our knowledge, we are the first to explore the use of DPO for aligning models with text quality preferences. 

\medskip
\noindent \textbf{\em Multilingual Evaluation} \quad 
Recently, \citet{sheng2024repeval} proposed an evaluation metric requiring LLMs to exhibit strong proficiency in each language being assessed, while \citet{trokhymovychetal2024open} measured readability scores using a ranking-based fine-tuning approach. \citet{hadaetal2024large} evaluated multiple dimensions across 10 languages. In our study, we aim to enhance the capability of models to effectively differentiate between high- and low-quality text in a multilingual setting, expanding language coverage to 115 languages.



\section{Conclusion}
We present a new framework to enhance open-source LLMs' capability to differentiate between high and low-quality text across multiple languages without the need for human annotations. Our approach demonstrates significant effectiveness across several datasets in various languages and also boosts downstream performance, that helps models to better grasp the distinction between good and poor-quality text. Measurable improvements on downstream tasks suggest that our approach does not just benefit evaluation but also enhances performance in other tasks, opening exciting possibilities.

\section*{Limitations}
This paper does not cover all languages, leaving several unexplored. While our text degradation methods are effective and scalable for simulating certain types of errors (e.g., grammatical disruptions, loss of coherence), they do not capture other dimensions of low-quality text, such as semantic errors, factual inaccuracies, or nuanced linguistic issues. Furthermore, applying the same degradation methods  uniformly across languages may not account for language-specific features. For instance, some languages (e.g., agglutinative or highly inflected languages) may not exhibit noticeable degradation from simple word shuffling, and future research should exploit the linguistic diversity of the dataset. Additionally, resource constraints prevented us from experimenting with larger models.

\section*{Ethics Statement}
We could  not examine whether the datasets used in this study contain  biases toward certain languages, potentially introduced by human annotators.

\section*{Acknowledgement}
We thank Dr. Suresh Singh, PortNLP Lab members, and the anonymous reviewers for their invaluable feedback.

\bibliography{custom}

\appendix
\section{Appendix}
\label{sec:appendix}


\subsection{Dataset format for DPO}
\label{sec:dpo_dataset_format}
Figures~\ref{fig:dpo_dataset_format_prompt} through \ref{fig:dpo_dataset_format_rejected} outline the prompts used in the dataset creation.

\begin{figure*}[!t]
\centering
\begin{tcolorbox}[colback=gray!10, colframe=gray, width=1.0\textwidth, rounded corners]

\begin{verbatim}
['Please act as an impartial judge and evaluate the text quality of the
provided passage.\n\nText quality of a passage is defined by how well it
maintains the following aspects.\n(1) Coherence - logical flow and 
connectivity between sentences and ideas in the text.\n(2) Fluency - 
smoothness and naturalness of individual sentences.\n(3) Simplicity - how
easy it is to understand the passage.\n(4) Linguistic Acceptability - if
the text sounds natural and correct to a native speaker.\n\nProvide a 
binary rating, “0” for low quality or “1" for high quality, strictly 
following this format: “[[0]]” or “[[1]]“. Do not provide an explanation.
\n\nPassage:\n\nL\'origami Pureland est un origami avec la contrainte qu
\'un seul plipeut être fait à la fois, les plis plus complexes comme les 
plis inversésne sont pas autorisés, et tous les plis ont des emplacements 
simples. Il a été développé par John Smith dans les années 1970 pour aider 
les personnes inexpérimentées ou ayant des capacités motrices limitées.']
\end{verbatim}
\end{tcolorbox}
\caption{An example of \emph{prompt} part of the DPO finetuning dataset}
\label{fig:dpo_dataset_format_prompt}
\end{figure*}

\begin{figure*}[!t]
\centering
\begin{tcolorbox}[colback=gray!10, colframe=gray, width=1.0\textwidth, rounded corners]

\begin{verbatim}
[[{'content': 'Please act as an impartial judge and evaluate the text 
quality of the provided passage.\n\nText quality of a passage is defined
by how well it maintains the following aspects.\n(1) Coherence - logical
flow and connectivity between sentences and ideas in the text.\n(2)
Fluency - smoothness and naturalness of individual sentences.\n(3)
Simplicity - how easy it is to understand the passage.\n(4) Linguistic
Acceptability - if the text sounds natural and correct to a native 
speaker.\n\nProvide a binary rating, “0” for low quality or “1" for high 
quality, strictly following this format: “[[0]]” or “[[1]]“. Do not 
provide an explanation.\n\nPassage:\n\nL\'origami Pureland est un origami 
avec la contrainte qu\'un seul pli peut être fait à la fois, les plis 
plus complexes comme les plis inversés ne sont pas autorisés, et tous 
les plis ont des emplacements simples. Il a été développé par John 
Smith dans les années 1970 pour aider les personnes inexpérimentées ou 
ayant des capacités motrices limitées.',
'role': 'user'},
{'content': 'Please act as an impartial judge and evaluate the text
quality of the provided passage.\n\nText quality of a passage is defined
by how well it maintains the following aspects.\n(1) Coherence - logical
flow and connectivity between sentences and ideas in the text.\n(2) 
Fluency - smoothness and naturalness of individual sentences.\n(3) 
Simplicity - how easy it is to understand the passage.\n(4) Linguistic
Acceptability - if the text sounds natural and correct to a native
speaker.\n\nProvide a binary rating, “0” for low quality or “1" for 
high quality, strictly following this format: “[[0]]” or “[[1]]“. Do 
not provide an explanation.\n\nPassage:\n\nL\'origami Pureland est un 
origami avec la contrainte qu\'un seul pli peut être fait à la fois, les 
plis plus complexes comme les plis inversés ne sont pas autorisés, et tous 
les plis ont des emplacements simples. Il a été développé par John Smith
dans les années 1970 pour aider les personnes inexpérimentées ou ayant
des capacités motrices limitées. \nRating: [[1]]',
'role': 'assistant'}]]
\end{verbatim}
\end{tcolorbox}
\caption{An example of \emph{chosen} part of the DPO finetuning dataset}
\label{fig:dpo_dataset_format_chosen}
\end{figure*}

\begin{figure*}[!t]
\centering
\begin{tcolorbox}[colback=gray!10, colframe=gray, width=1.0\textwidth, rounded corners]

\begin{verbatim}
[[{'content': 'Please act as an impartial judge and evaluate the text 
quality of the provided passage.\n\nText quality of a passage is defined
by how well it maintains the following aspects.\n(1) Coherence - logical
flow and connectivity between sentences and ideas in the text.\n(2)
Fluency - smoothness and naturalness of individual sentences.\n(3)
Simplicity - how easy it is to understand the passage.\n(4) Linguistic
Acceptability - if the text sounds natural and correct to a native 
speaker.\n\nProvide a binary rating, “0” for low quality or “1" for high 
quality, strictly following this format: “[[0]]” or “[[1]]“. Do not 
provide an explanation.\n\nPassage:\n\nL\'origami Pureland est un origami 
avec la contrainte qu\'un seul pli peut être fait à la fois, les plis 
plus complexes comme les plis inversés ne sont pas autorisés, et tous les
plis ont des emplacements simples. Il a été développé par John Smith 
dans les années 1970 pour aider les personnes inexpérimentées ou ayant
des capacités motrices limitées.',
   'role': 'user'},
{'content': 'Please act as an impartial judge and evaluate the text
quality of the provided passage.\n\nText quality of a passage is defined
by how well it maintains the following aspects.\n(1) Coherence - logical
flow and connectivity between sentences and ideas in the text.\n(2) 
Fluency - smoothness and naturalness of individual sentences.\n(3) 
Simplicity - how easy it is to understand the passage.\n(4) Linguistic
Acceptability - if the text sounds natural and correct to a native
speaker.\n\nProvide a binary rating, “0” for low quality or “1" for high 
quality, strictly following this format: “[[0]]” or “[[1]]“. Do not 
provide an explanation.\n\nPassage:\n\nL\'origami Pureland est un origami 
avec la contrainte qu\'un seul pli peut être fait à la fois, les plis 
plus complexes comme les plis inversés ne sont pas autorisés, et tous les
plis ont des emplacements simples. Il a été développé par John Smith
dans les années 1970 pour aider les personnes inexpérimentées ou ayant
des capacités motrices limitées. \nRating: [[0]]',
'role': 'assistant'}]]
\end{verbatim}
\end{tcolorbox}
\caption{An example of \emph{rejected} part of the DPO finetuning dataset}
\label{fig:dpo_dataset_format_rejected}
\end{figure*}

\subsection{Prompts used in different scenarios}
\label{sec:prompt_dimension_specific}

Figures \ref{fig:mela_prompt} through
\ref{fig:xlsum_sum_prompt} outline the various prompts used in this study.

\subsection{Full results of top and bottom performing languages}\label{app:top}

Tables~\ref{app:f1_top_bottom_llama_ayaexp} presents the full results of top and bottom performing languages on Belebele text quality evaluation dataset.

\begin{figure*}[!t]
\centering
\begin{tcolorbox}[colback=gray!10, colframe=gray, width=1.0\textwidth, rounded corners]

\begin{verbatim}
Determine whether the following sentence(s) violate certain linguistic 
constraints. If yes, then it is "unacceptable"; otherwise, "acceptable".

Sentence: {sentence}

Determine whether this sentence is acceptable or unacceptable? If 
acceptable, return [[1]], otherwise [[0]]. Strictly follow this format: 
“[[0]]” or “[[1]]“. Do not provide any feedback.

Your Answer:
\end{verbatim}
\end{tcolorbox}
\caption{Prompt used for MELA evaluation.}
\label{fig:mela_prompt}
\end{figure*}

\begin{figure*}[!t]
\centering
\begin{tcolorbox}[colback=gray!10, colframe=gray, width=1.0\textwidth, rounded corners]

\begin{verbatim}
Please act as an impartial judge and determine the sentiment of the 
following passage.

Passage: {text}

If the sentiment is negative, return [[0]], if neutral, return [[1]], and 
if positive, return [[2]]. Strictly follow this format: “[[0]]”, “[[1]]”, 
or “[[2]]“. Do not provide any feedback.

Your Answer:
\end{verbatim}
\end{tcolorbox}
\caption{Prompt used for MMS evaluation.}
\label{fig:mms_prompt}
\end{figure*}

\begin{figure*}[!t]
\centering
\begin{tcolorbox}[colback=gray!10, colframe=gray, width=1.0\textwidth, rounded corners]

\begin{verbatim}
Summarize the following article in one or two sentence. 

Article:
{text}

Do not include any additional note or text; simply output one sentence 
summary and nothing more.

Your Answer:
\end{verbatim}
\end{tcolorbox}
\caption{Prompt used for XLSum summary generation.}
\label{fig:xlsum_sum_prompt}
\end{figure*}

\begin{table*}[!t]\centering
\setlength{\tabcolsep}{6pt}
\small
\begin{tabular}{lclc|lclc}
\toprule
\multicolumn{4}{c}{\textbf{MTQ-Eval \texttt{llama}}} & \multicolumn{4}{c}{\textbf{MTQ-Eval \texttt{aya}}}\\
\textbf{Language} &\textbf{F1 (Top 20)} &\textbf{Language} &\textbf{F1 (Bottom 20)} &\textbf{Language} &\textbf{F1 (Top 20)} &\textbf{Language} &\textbf{F1 (Bottom 20)} \\
\midrule
eng\_Latn &0.95 &sin\_Sinh &0.44 &spa\_Latn &0.97 &tso\_Latn &0.33 \\
nld\_Latn &0.80 &guj\_Gujr &0.44 &pol\_Latn &0.97 &tsn\_Latn &0.33 \\
ron\_Latn &0.80 &npi\_Deva &0.42 &eng\_Latn &0.97 &yor\_Latn &0.33 \\
nob\_Latn &0.80 &hau\_Latn &0.42 &fra\_Latn &0.95 &xho\_Latn &0.33 \\
pes\_Arab &0.77 &kac\_Latn &0.42 &ell\_Grek &0.95 &ory\_Orya &0.33 \\
por\_Latn &0.77 &tir\_Ethi &0.40 &deu\_Latn &0.95 &grn\_Latn &0.33 \\
swe\_Latn &0.76 &fuv\_Latn &0.40 &ces\_Latn &0.92 &hat\_Latn &0.33 \\
nya\_Latn &0.75 &kat\_Geor &0.39 &ron\_Latn &0.92 &hau\_Latn &0.33 \\
zsm\_Latn &0.74 &lao\_Laoo &0.39 &ita\_Latn &0.92 &ibo\_Latn &0.33 \\
ita\_Latn &0.74 &urd\_Arab &0.39 &nld\_Latn &0.92 &ssw\_Latn &0.33 \\
fra\_Latn &0.73 &mal\_Mlym &0.39 &ind\_Latn &0.9 &kac\_Latn &0.33 \\
spa\_Latn &0.73 &khm\_Khmr &0.39 &por\_Latn &0.9 &sot\_Latn &0.33 \\
isl\_Latn &0.72 &kan\_Knda &0.37 &ukr\_Cyrl &0.87 &sna\_Latn &0.33 \\
hat\_Latn &0.72 &bod\_Tibt &0.37 &pes\_Arab &0.87 &kin\_Latn &0.33 \\
ilo\_Latn &0.72 &amh\_Ethi &0.37 &vie\_Latn &0.87 &sin\_Sinh &0.33 \\
als\_Latn &0.70 &asm\_Beng &0.37 &arb\_Arab &0.87 &lin\_Latn &0.33 \\
uzn\_Latn &0.70 &tel\_Telu &0.36 &cat\_Latn &0.85 &lug\_Latn &0.33 \\
jav\_Latn &0.70 &ben\_Beng &0.36 &slk\_Latn &0.82 &nso\_Latn &0.33 \\
slk\_Latn &0.69 &ory\_Orya &0.33 &lit\_Latn &0.80 &pbt\_Arab &0.33 \\
azj\_Latn &0.69 &tam\_Taml &0.33 &heb\_Hebr &0.80 &zul\_Latn &0.33 \\
\bottomrule
\end{tabular}
\caption{Top and bottom (20) performing languages on Belebele text quality evaluation dataset.}
\label{app:f1_top_bottom_llama_ayaexp}
\end{table*}

\end{document}